\documentclass[11pt]{article}

\usepackage[preprint]{acl}
\usepackage[most]{tcolorbox} 
\usepackage{times}
\usepackage{latexsym}
\usepackage[T1]{fontenc}
\usepackage[utf8]{inputenc}
\usepackage{microtype}
\usepackage{inconsolata}
\usepackage{graphicx}
\usepackage{amsmath,amssymb}
\usepackage{booktabs}
\usepackage{array}
\usepackage{multirow}
\usepackage[table]{xcolor}
\definecolor{somabg}{RGB}{232,240,252}
\usepackage{tabularx}
\usepackage{algorithm}
\usepackage{algpseudocode}
\usepackage{placeins}
\usepackage{enumitem}
\expandafter\def\expandafter\UrlBreaks\expandafter{\UrlBreaks\do\-}
\newcommand{\soma}{SOMA}

\title{SOMA: Continuous Monitoring and Intervention for System Two Reasoning with System One Models}

\author{
  Ruizhi Qian \quad Lei Shi \quad Tianyu Hu \quad Bo Yan \quad Song Wang \quad Yu Tian \\
  University of Central Florida \\
  {\small\texttt{ru282875@ucf.edu, le779457@ucf.edu, ti721494@ucf.edu}} \\
  {\small\texttt{bo949643@ucf.edu, song.wang@ucf.edu, yu.tian2@ucf.edu}}
}

\begin{document}
\maketitle

\begin{abstract}
Reasoning models often run into problems without noticing them, even when symptoms are visible in their own text. In humans, metacognition separates monitoring from control: fast, intuitive signals keep watch while slow, deliberate reasoning unfolds and help decide whether to continue, change course, or stop. System One (S1) decision models, such as Jev, are a natural candidate for this role: they return a calibrated probability for a natural-language question about a text in one fast call, yet they are used only as gates before reasoning. We introduce SOMA (System One Metacognitive Awareness), in which an S1 model continuously monitors a System Two (S2) reasoning model and intervenes when needed. SOMA has three layers: a scenario layer states each objective in natural language with predefined actions, an awareness layer lets S1 judge whether the described behavior appears, and a control layer accumulates these judgments, first reminds S2, and takes over only if the behavior persists. Sharing one S1 model and one controller across tasks, without training, SOMA addresses two different failures on two reasoning models: on overthinking, it cuts reasoning tokens by up to 36\% while improving average accuracy, and on unanswerable questions, it raises abstention from 70.5\% to 87.2\%.

% Reasoning models often run into problems during long chains of thought without noticing them, even when the symptoms are visible in their own text. In humans, metacognition separates monitoring from control: fast, intuitive signals keep watch while slow, deliberate reasoning unfolds, and those signals help decide whether to continue, change course, or stop. Recent System One (S1) decision models, such as Jev, are a natural candidate for this monitoring role: they return a calibrated probability for a natural-language question about a text in one fast call, yet they are used only as gates before reasoning begins. Building on this, we introduce SOMA (System One Metacognitive Awareness), in which an S1 model continuously monitors a System Two (S2) reasoning model and intervenes when needed. SOMA has three layers: a scenario layer states each objective in natural language with predefined actions, an awareness layer lets S1 judge whether the described behavior appears, and a control layer accumulates these judgments, first reminds S2, and takes over only if the behavior persists. Because S1 reads only the visible reasoning, any reasoning model can be supervised, and a new objective needs only a new statement. Sharing one S1 model and one controller across tasks, without any training, SOMA addresses two different failures on two reasoning models: on overthinking, it cuts reasoning tokens by up to 36\% while improving average accuracy, and on unanswerable questions, it raises abstention from 70.5\% to 87.2\%.
\end{abstract}

\section{Introduction}
\label{sec:intro}

Reasoning models such as DeepSeek-R1 and Qwen3 solve problems through long chains of thought \citep{guo2025deepseek,yang2025qwen3}, yet they often run into problems without noticing them, even when the symptoms are visible in their own text. They keep re-verifying an answer they have already confirmed \citep{chen2025think,peng2025revisiting}, circle back to the same doubts, or state that a problem is missing information and still force a definite answer \citep{fan2025missing,liu2026answering}. The trace already shows what has gone wrong, but what is lacking is that nothing notices it while the reasoning is still under way.

Human reasoning suggests such an observer. Research on metacognition distinguishes the \emph{monitoring} of one's own thinking from its \emph{control} \citep{flavell1979metacognition,nelson1990metamemory}: fast, intuitive signals, such as a feeling of rightness or a sensed conflict, accompany slow deliberation and help decide whether it continues, changes course, or stops \citep{thompson2011intuition,deneys2008conflict,ackerman2017meta}.

\begin{figure*}[t]
\centering
\includegraphics[width=\textwidth]{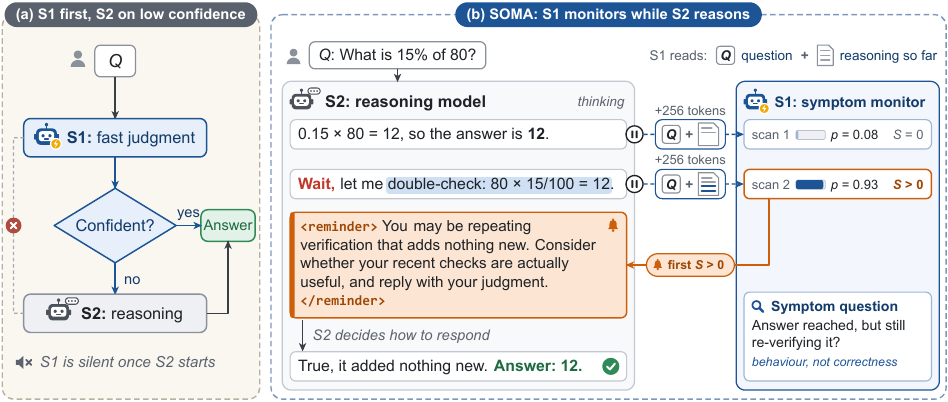}
\caption{(a) In a common combination, S1 decides first and S2 is called for low-confidence cases; S1 is silent once S2 starts. (b) In \soma{}, S1 runs beside S2: S2 periodically pauses at a line break, S1 judges whether a statement about an observable behavior holds for the reasoning so far, and cumulative evidence from positive judgments triggers a reminder inserted at the pause, where S2 decides how to respond. The trace is constructed for illustration, not taken from a run; the evidence accumulation and takeover rules are in Section~\ref{sec:control}.}
\label{fig:paradigm}
\end{figure*}

Recent System One decision models are a natural candidate for this monitoring role. Exemplified by Jev \citep{almeida2026introducing}, they take a text and a natural-language question about it and return a calibrated probability in a single fast call \citep{typesafe2026documentation}. Following dual-process accounts \citep{kahneman2011thinking,evans2013dual}, we call these models System One (S1) and reasoning models System Two (S2), as functional labels for fast judgment and slow deliberation.

Two limitations of this design motivate our work. First, S1 models are used only as gates before reasoning begins. Jev's documentation, for instance, includes an example that uses it to escalate doubtful cases to a reasoning model \citep{typesafe2026documentation} (Figure~\ref{fig:paradigm}(a)). Once S2 starts, S1 falls silent, and a trace of tens of thousands of tokens unfolds without a fast monitor. Second, existing remedies for these reasoning problems are tailored to one scenario at a time: each ties a dedicated signal, such as trial-answer confidence, answer agreement or a hidden-state probe, to a single action \citep{yang2026dynamic,fu2025efficiently,liu2026answering,xie2026statistical}, and many require training or white-box access to the model \citep{aggarwal2025l1,zhang2025reasoning,liu2026answering}. Each new problem thus calls for a new mechanism.

To address both gaps, we introduce \soma{} (System One Metacognitive Awareness), in which an S1 model monitors S2 throughout generation and intervenes when needed (Figure~\ref{fig:paradigm}(b)). \soma{} has three layers. The \emph{scenario} layer states an objective in natural language and lists the actions allowed. The \emph{awareness} layer asks S1 whether that behavior appears in the trace. The \emph{control} layer accumulates these judgments, reminds S2 first, and takes over only if the behavior persists. We instantiate \soma{} for overthinking and unanswerable questions, which share the same S1 model, controller and system prompt and differ only in their monitors. Our contributions are as follows:
% \begin{itemize}
\begin{itemize}[leftmargin=*, nosep]
\item We propose using S1 decision models as continuous monitors of S2 reasoning rather than gates before it, and introduce \soma{}, which separates supervision into scenario, awareness and control layers so that a new objective requires only a natural-language specification.
\item We design a remind-first, take-over-on-persistent-evidence controller with a cumulative log-odds rule, and a calibration procedure that limits the harm of irreversible takeovers.
\item With a single black-box S1 and no training, \soma{} reduces reasoning tokens on overthinking by 36\% and 16\% without lowering average accuracy, and raises abstention on unanswerable questions by 17 and 21 points on two reasoning models.
\end{itemize}

\section{Related Work}
\label{sec:related}

\subsection{Dual-Process Reasoning and Metacognition}
Dual-process theories distinguish fast, intuitive judgments from slow, deliberate reasoning \citep{sloman1996empirical,kahneman2011thinking,evans2013dual}. Metacognitive accounts further distinguish monitoring from control: signals such as confidence and detected conflict can determine whether deliberation continues, changes course, or stops \citep{flavell1979metacognition,nelson1990metamemory,ackerman2017meta,deneys2008conflict,thompson2011intuition,ackerman2014diminishing}. We use this relationship as a functional analogy for an S1 model that observes an S2 reasoning trace and supplies judgments to a controller.

\subsection{System One Decision Models}
This distinction has recently reappeared in model design. System One decision models return typed, probabilistic judgments instead of free-form text. Jev was an early public example; related models now include CLM, Laya, OpenAI's Decisions API (\texttt{gpt-6-luna}), and Microsoft-Decision-1 \citep{almeida2026introducing,typesafe2026documentation,contrastivelm2026clm,convai2026laya,openai2026decisions,microsoft2026decision}. Their common interface makes a fast judgment available to an application or controller without asking a generative model to write a rationale, although their probabilities still require careful calibration and thresholding \citep{deusser2026evaluating,porcedda2026jev,rafe2026benchmarking}, as do earlier fast judges and guard models \citep{kadavath2022language,zheng2023judging,inan2023llama}. Like other fast models, they are mostly used before generation: cascades and routers escalate uncertain inputs, and fast-and-slow architectures invoke a stronger solver when the fast one falls short \citep{chen2024frugalgpt,aggarwal2024automix,ong2025routellm,lin2023swiftsage,ganapini2025fast,khandelwal2025language}.

\subsection{Monitoring Reasoning During Generation}
A few methods supervise a reasoning model while it generates. TrimR and FlashThink query a separate prompted LLM at reasoning boundaries, asking whether a conclusion has been reached or whether the thought so far suffices, and cut thinking short when it does \citep{lin2026trimr,jiang2025flashthink}. Meta-R1 and MetaCtrl let a meta-level component steer the reasoner, using a generative language model or a controller trained for concise reasoning \citep{dong2025meta,wen2026metactrl}. Each of these methods targets a single objective with a fixed question and action. \soma{} instead pairs an S1 decision model with a shared controller, so that different objectives share one interface.

\subsection{Inference-Time Control of Reasoning}
We study two cases in which reasoning should change course after generation has begun. In overthinking, a model keeps revisiting an answer after reaching a useful conclusion \citep{chen2025think,peng2025revisiting,sui2025stop}; existing methods shorten reasoning by training the model or by stopping on trial-answer confidence, answer agreement, or hidden-state probes \citep{aggarwal2025l1,zhang2025adaptthink,fang2025thinkless,yang2026dynamic,fu2025efficiently,zhang2025reasoning,wang2026rom}. In unanswerable-question settings, models may continue reasoning despite missing premises or an explicitly noticed flaw \citep{kirichenko2025abstentionbench,song2025hallucination,fan2025missing,ahn2026bridging,liu2026answering}; prior remedies use a hidden-state probe to trigger abstention guidance \citep{liu2026answering} or sequential statistical tests on uncertainty keywords to stop reasoning \citep{xie2026statistical}. These methods generally couple a scenario to a dedicated signal and action. \soma{} instead expresses both cases as monitor specifications and reuses the same S1 model and controller, differing only in the behavior statement and predefined response.

\section{SOMA}
\label{sec:method}

\begin{figure*}[t]
\centering
\includegraphics[width=\textwidth]{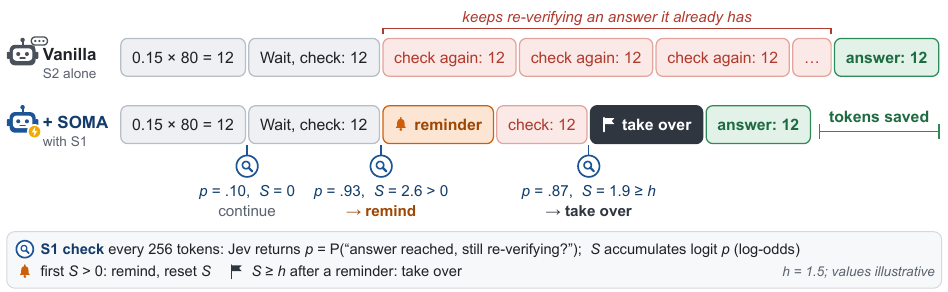}
\caption{Overview of \soma{}. Every 256 thinking tokens, S2 continues to the next line break (at most 128 more tokens) and pauses. S1 reads the task, the visible reasoning, and a monitor statement, and returns a probability $p$, which updates the cumulative score $S$. Background shading shows the controller state: the first positive $S$ inserts a reminder and resets $S$ (orange); post-reminder evidence reaching the frozen level $h$ triggers a setting-specific takeover (grey), after which S2 writes a bounded answer or refusal reason. The trace is illustrative.}
\label{fig:overview}
\end{figure*}

\soma{} lets an S1 decision model supervise an S2 reasoning model while S2 is thinking. Following the separation of monitoring and control in human metacognition, it divides supervision into a \emph{scenario layer} that defines what to supervise, an \emph{awareness layer} in which S1 detects it, and a \emph{control layer} that decides how to respond with a single rule shared by all scenarios (Figure~\ref{fig:overview}). Each layer can be replaced on its own: a new objective only adds a scenario specification, and a different S1 model plugs in through the same interface.

Throughout this paper, $Q$ denotes the task and $H_t$ the reasoning text that S2 has written after $t$ thinking tokens.

\subsection{Scenario Layer: What to Supervise}
\label{sec:scenario}

A supervision objective is specified by a \emph{monitor} $m=(s_m,r_m,a_m)$, consisting entirely of natural-language and predefined text.

\paragraph{Statement $s_m$.} The statement describes the reasoning behavior to watch for, and it is the only part of the monitor that S1 sees. Statements follow a common pattern: \emph{X has appeared in the reasoning, but S2 is still doing Y}. This form has two advantages. Both parts can be read off the visible text, unlike the correctness of an answer, and the statement stops being true once S2 changes its behavior, so S1's later judgments show whether an intervention has taken effect.

\paragraph{Reminder $r_m$.} The reminder is a short message inserted into S2's reasoning. Rather than issuing a command, it points to the possible problem and asks S2 to judge whether its current activity is useful, leaving the decision to S2.

\paragraph{Takeover action $a_m$.} The takeover action is predefined text that ends thinking and steers the response, for example towards a final answer or a refusal. S2 still writes the final content after a takeover.

Because a monitor is plain text, adding a scenario requires neither training nor any change to the other two layers. The same form could also describe a reasoning trace that starts to comply with a jailbreak request.

\subsection{Awareness Layer: S1 as an Observer}
\label{sec:awareness}

S1 is queried at regular pauses while S2 is thinking. At each pause, it receives the task, the reasoning so far and the statement, and returns
\[
p_t=\mathrm{S1}(Q,H_t,s_m)\in[0,1],
\]
the probability that the statement holds. This interface is the only requirement on S1: any decision model that maps a text state and a natural-language statement to a probability can fill the role. It also sees only text written by S2. Inserted reminders and the system prompt are hidden from it, so its judgment always concerns S2's own behavior.

Pauses occur every $X$ thinking tokens. To avoid reading a half-written line, S2 first continues to the next line break and pauses there. Because S1 returns a single probability rather than generated text, each scan is one short call. When the reasoning exceeds S1's input limit, the task is kept in full and the earliest part of the reasoning is dropped.

\subsection{Control Layer: Remind First, Take Over on Persistent Evidence}
\label{sec:control}

The controller turns S1's judgments into evidence on the log-odds scale and accumulates it as a one-sided cumulative sum:
\[
S\leftarrow\max\bigl\{0,\;S+\operatorname{logit}(p_t)-\operatorname{logit}(0.5)\bigr\}.
\]
A scan with $p_t>0.5$ raises $S$, a scan with $p_t<0.5$ lowers it toward zero, and a scan with $p_t=0.5$ leaves it unchanged; $S$ never drops below zero.

This rule follows the one-sided cumulative-sum test for detecting a persistent change \citep{page1954continuous}. Since S1 reports the probability that the statement holds, $\operatorname{logit}(p_t)-\operatorname{logit}(0.5)$ is the evidence of one scan for the statement against even odds, and summing log-odds treats successive scans as separate pieces of evidence. A confident scan therefore counts more than a doubtful one, an isolated high score is absorbed by later low ones, and only a behavior that S1 keeps observing drives $S$ upward.

The controller responds in two stages (Algorithm~\ref{alg:soma}). When $S$ first becomes positive, an \emph{episode} begins: \soma{} inserts the reminder $r_m$ at the pause and resets $S$ to zero. S2 then continues and decides for itself how to respond. From this point $S$ accumulates only evidence gathered after the reminder. If it reaches a threshold $h$, the controller executes the takeover action $a_m$; if it falls back to zero first, the episode ends, and a later rise of $S$ above zero starts a new one with a new reminder.

The two stages differ in cost and reversibility. A reminder leaves S2 in charge and costs only a few tokens, whereas a takeover cannot be undone by S2. \soma{} therefore takes over only when the behavior persists after S2 has been given a chance to stop, and it bases this decision on what S1 observes S2 doing after the reminder, not on what S2 says in reply. To make S2 attend to reminders, a single fixed system prompt, shared by all scenarios, tells S2 that a monitor may insert reminders into its reasoning.

\begin{algorithm}[t]
\footnotesize
\caption{\soma{} supervision loop for monitor $m$.}
\label{alg:soma}
\begin{algorithmic}[1]
\Require task $Q$, S2, S1, monitor $m=(s_m,r_m,a_m)$, threshold $h$
\State $S\gets 0$;\; $\mathit{reminded}\gets\mathit{false}$
\While{S2 is thinking}
  \State run S2 to the next pause; $H\gets$ its reasoning
  \State $p\gets\mathrm{S1}(Q,H,s_m)$;\; $S\gets\max\{0,\,S+\operatorname{logit}(p)-\operatorname{logit}(0.5)\}$
  \If{$S=0$}
    \State $\mathit{reminded}\gets\mathit{false}$ \Comment{episode ends}
  \ElsIf{\textbf{not} $\mathit{reminded}$}
    \State insert $r_m$;\; $S\gets 0$;\; $\mathit{reminded}\gets\mathit{true}$
  \ElsIf{$S\ge h$}
    \State apply $a_m$ and stop thinking \Comment{takeover}
  \EndIf
\EndWhile
\end{algorithmic}
\end{algorithm}

\subsection{Threshold Calibration}
\label{sec:calibration}

A takeover is the one intervention S2 cannot recover from. Once thinking is cut off and the response is steered, a mistaken takeover becomes a wrong answer or a false refusal, whereas a mistaken reminder can simply be dismissed by S2. The threshold $h$ therefore decides how much post-reminder evidence \soma{} requires before it accepts this risk.

No single threshold suits every setting: the evidence that $S$ accumulates depends on how sharply S1 separates cases in which the statement holds from those in which it does not \citep{deusser2026evaluating}, on how S2 responds to reminders, and on how often a monitor's behavior occurs and what a mistaken takeover costs. \soma{} therefore calibrates a separate threshold for each combination of S1, S2 and monitor.

Calibration takes place before deployment, on a small set $\mathcal{C}$ of representative tasks for the target scenario. On these tasks, \soma{} under each candidate threshold $h\in\mathcal{H}$ is compared with S2 running under the same system prompt without any intervention, so that every difference between the two is caused by the intervention. Let $c_0(q)\in\{0,1\}$ indicate whether S2 handles task $q$ correctly on its own, and $c_h(q)$ whether it does so under \soma{} with threshold $h$. In overthinking, a task is handled correctly when the final answer is correct; for unanswerable questions, when S2 abstains on an unanswerable question or correctly answers an answerable one. Harm thus covers both answers spoiled by an early takeover and false refusals. We measure the harm of a threshold as the number of tasks that the intervention turns from correct to wrong:
\[
L(h)=\sum_{q\in\mathcal{C}}c_0(q)\,\bigl(1-c_h(q)\bigr),
\]
and choose the smallest threshold whose harm is within a tolerance $\delta\ge 0$ of the lowest achievable harm:
\[
h^{*}=\min\Bigl\{h\in\mathcal{H}:\ L(h)\le\min_{h'\in\mathcal{H}}L(h')+\delta\Bigr\}.
\]
A smaller threshold takes over earlier and saves more computation, so the rule selects the most aggressive threshold whose harm stays within the tolerance; $\delta$ sets how much harm is acceptable in exchange for savings. Once chosen, the threshold stays fixed during use.

\section{Experiments}
\label{sec:experiments}

\begin{table*}[t]
\centering
\small
\renewcommand{\arraystretch}{1.15}
\begin{tabularx}{\textwidth}{@{}>{\raggedright\arraybackslash}p{1.85cm}>{\raggedright\arraybackslash}X>{\raggedright\arraybackslash}X>{\raggedright\arraybackslash}p{3.6cm}@{}}
\toprule
\textbf{Scenario} & \textbf{Statement $s_m$ (judged by S1)} & \textbf{Reminder $r_m$} & \textbf{Takeover action $a_m$} \\
\midrule
Overthinking &
The reasoning has reached a final answer and keeps re-verifying it without naming any specific remaining concern. &
You may be repeating verification that adds nothing new. Consider whether your recent checks are actually useful, and reply with your judgment. &
End thinking; insert \texttt{**Final Answer**}; S2 writes the answer (64 tokens). \\
\addlinespace
Unanswerable questions &
The reasoning has noticed that the problem is missing necessary information or has contradictory conditions, but is still trying to produce a definite answer, for example by assuming values the problem does not give. &
Check once more whether the problem gives everything you need. Then judge whether this check was useful or merely repeated your earlier checks, and say so explicitly. &
Insert ``The problem is missing necessary information or has contradictory conditions, so it cannot be answered as stated.''; end thinking; pre-fill \texttt{\textbackslash boxed\{I don't know.\}}; S2 writes the reason (512 tokens). \\
\bottomrule
\end{tabularx}
\caption{The two monitors evaluated in this paper. Each reminder is wrapped in \texttt{<reminder>} tags and inserted at a pause. Both scenarios share the S1 model, the controller and the system prompt.}
\label{tab:monitors}
\end{table*}

\begin{table*}[t]
\centering\small
\setlength{\tabcolsep}{3.6pt}
\begin{tabular}{@{}lrrrrrrrrrrrr@{}}
\toprule
 & \multicolumn{2}{c}{GSM8K} & \multicolumn{2}{c}{MATH-500} & \multicolumn{2}{c}{AIME} & \multicolumn{2}{c}{AMC23} & \multicolumn{2}{c}{GPQA-D} & \multicolumn{2}{c}{\textbf{Avg}} \\
\cmidrule(lr){2-3}\cmidrule(lr){4-5}\cmidrule(lr){6-7}\cmidrule(lr){8-9}\cmidrule(lr){10-11}\cmidrule(lr){12-13}
Method & Acc$\uparrow$ & Tok & Acc$\uparrow$ & Tok & Acc$\uparrow$ & Tok & Acc$\uparrow$ & Tok & Acc$\uparrow$ & Tok & Acc$\uparrow$ & Tok \\
\midrule
\multicolumn{13}{@{}l}{\textit{Qwen3-8B}} \\
Vanilla & 95.2 & 2,277 & \textbf{96.2} & 5,549 & 66.7 & 16,719 & \textbf{95.0} & 8,600 & \underline{56.6} & 9,266 & 91.0 & 4,232 \\
BF & \underline{95.3} & {\scriptsize $-$1\%} & 95.4 & {\scriptsize $-$9\%} & 55.0 & {\scriptsize\underline{$-$32\%}} & \textbf{95.0} & {\scriptsize $-$10\%} & 55.1 & {\scriptsize $-$7\%} & 90.4 & {\scriptsize $-$8\%} \\
DEER & 94.2 & {\scriptsize\boldmath\textbf{$-$62\%}} & 92.4 & {\scriptsize\boldmath\textbf{$-$52\%}} & 55.0 & {\scriptsize\boldmath\textbf{$-$36\%}} & 77.5 & {\scriptsize\boldmath\textbf{$-$56\%}} & 49.0 & {\scriptsize\boldmath\textbf{$-$53\%}} & 88.1 & {\scriptsize\boldmath\textbf{$-$54\%}} \\
Dynasor-CoT & \underline{95.3} & {\scriptsize $-$3\%} & \underline{95.8} & {\scriptsize $-$6\%} & \underline{68.3} & {\scriptsize $-$7\%} & \textbf{95.0} & {\scriptsize $-$6\%} & \textbf{60.1} & {\scriptsize $-$12\%} & \textbf{91.4} & {\scriptsize $-$6\%} \\
\rowcolor{somabg}\soma{} (ours) & \textbf{95.5} & {\scriptsize\underline{$-$42\%}} & 94.8 & {\scriptsize\underline{$-$46\%}} & \textbf{73.3} & {\scriptsize $-$29\%} & \underline{90.0} & {\scriptsize\underline{$-$37\%}} & \textbf{60.1} & {\scriptsize\underline{$-$17\%}} & \underline{91.3} & {\scriptsize\underline{$-$36\%}} \\
\midrule
\multicolumn{13}{@{}l}{\textit{R1-Distill-Llama-8B}} \\
Vanilla & \textbf{91.2} & 1,747 & 90.8 & 4,289 & 36.7 & 13,718 & \underline{87.5} & 7,030 & 36.9 & 8,153 & 84.4 & 3,386 \\
BF & \textbf{91.2} & {\scriptsize $-$2\%} & \textbf{91.6} & {\scriptsize $-$7\%} & \underline{38.3} & {\scriptsize\boldmath\textbf{$-$14\%}} & \underline{87.5} & {\scriptsize $-$7\%} & 36.4 & {\scriptsize $-$0\%} & 84.6 & {\scriptsize $-$5\%} \\
DEER & 82.9 & {\scriptsize\boldmath\textbf{$-$40\%}} & 87.4 & {\scriptsize\boldmath\textbf{$-$30\%}} & \underline{38.3} & {\scriptsize\underline{$-$9\%}} & 82.5 & {\scriptsize\underline{$-$20\%}} & 33.8 & {\scriptsize\underline{$-$10\%}} & 78.1 & {\scriptsize\boldmath\textbf{$-$26\%}} \\
Dynasor-CoT & \textbf{91.2} & {\scriptsize $-$1\%} & \underline{91.0} & {\scriptsize $-$4\%} & 36.7 & {\scriptsize $-$5\%} & 85.0 & {\scriptsize $-$14\%} & \textbf{42.9} & {\scriptsize\boldmath\textbf{$-$45\%}} & \textbf{85.0} & {\scriptsize $-$13\%} \\
\rowcolor{somabg}\soma{} (ours) & \underline{90.8} & {\scriptsize\underline{$-$25\%}} & 90.4 & {\scriptsize\underline{$-$20\%}} & \textbf{45.0} & {\scriptsize $-$1\%} & \textbf{92.5} & {\scriptsize\boldmath\textbf{$-$24\%}} & \underline{41.4} & {\scriptsize $-$5\%} & \underline{84.8} & {\scriptsize\underline{$-$16\%}} \\
\bottomrule
\end{tabular}
\caption{Overthinking results. Acc is accuracy (\%); Tok is the mean number of S2 tokens, given in absolute terms for Vanilla and as the change relative to Vanilla otherwise. For \soma{}, the accounting follows the readout in Appendix~\ref{app:readout}. Avg weights each benchmark by its number of questions and is computed from unrounded per-benchmark values. Bold marks the best accuracy and the largest token reduction for each S2 model; underline marks the second best.}
\label{tab:overthinking}
\end{table*}

\begin{table*}[t]
\centering\small
\setlength{\tabcolsep}{3.4pt}
\begin{tabular}{@{}lrrrrrrrrrrrr@{}}
\toprule
 & \multicolumn{4}{c}{SUM} & \multicolumn{4}{c}{UMWP} & \multicolumn{4}{c}{\textbf{Avg}} \\
\cmidrule(lr){2-5}\cmidrule(lr){6-9}\cmidrule(lr){10-13}
Method & Abs$\uparrow$ & Acc$\uparrow$ & FA$\downarrow$ & Tok & Abs$\uparrow$ & Acc$\uparrow$ & FA$\downarrow$ & Tok & Abs$\uparrow$ & Acc$\uparrow$ & FA$\downarrow$ & Tok \\
\midrule
\multicolumn{13}{@{}l}{\textit{Qwen3-8B}} \\
Vanilla & 60 & \textbf{80} & \textbf{3} & 6,171 & 81 & \underline{97} & \textbf{2} & 1,627 & 70.5 & \textbf{88.7} & \textbf{2.4} & 3,837 \\
DEER & 68 & 73 & \underline{6} & {\scriptsize\boldmath\textbf{$-$53\%}} & 81 & 95 & \underline{3} & {\scriptsize\boldmath\textbf{$-$64\%}} & 74.8 & 84.2 & 4.6 & {\scriptsize\boldmath\textbf{$-$56\%}} \\
Dynasor-CoT & 61 & \underline{77} & \underline{6} & {\scriptsize $-$12\%} & 83 & \textbf{98} & \textbf{2} & {\scriptsize $-$15\%} & 72.1 & \underline{87.8} & \underline{3.8} & {\scriptsize $-$12\%} \\
AbsBench prompt & 79 & 65 & 17 & {\scriptsize +2\%} & 85 & 95 & \underline{3} & {\scriptsize $-$15\%} & 82.2 & 80.1 & 10.1 & {\scriptsize $-$2\%} \\
SES & 80 & \underline{77} & 10 & {\scriptsize $-$23\%} & \textbf{91} & 91 & 8 & {\scriptsize $-$35\%} & 85.8 & 83.9 & 8.9 & {\scriptsize $-$26\%} \\
AtU & \underline{83} & \underline{77} & \underline{6} & {\scriptsize\underline{$-$40\%}} & \underline{90} & 95 & 4 & {\scriptsize\underline{$-$41\%}} & \underline{86.5} & 86.6 & 5.0 & {\scriptsize\underline{$-$41\%}} \\
\rowcolor{somabg}\soma{} (ours) & \textbf{87} & \underline{77} & 11 & {\scriptsize $-$32\%} & 88 & 92 & 6 & {\scriptsize $-$31\%} & \textbf{87.2} & 84.9 & 8.0 & {\scriptsize $-$32\%} \\
\midrule
\multicolumn{13}{@{}l}{\textit{R1-Distill-Llama-8B}} \\
Vanilla & 28 & \underline{78} & \textbf{1} & 6,716 & 48 & \underline{86} & \textbf{3} & 1,893 & 38.2 & \underline{82.4} & \textbf{2.1} & 4,238 \\
DEER & 34 & 77 & \underline{2} & {\scriptsize $-$14\%} & 52 & \underline{86} & \textbf{3} & {\scriptsize $-$9\%} & 43.0 & 81.7 & \underline{2.9} & {\scriptsize $-$12\%} \\
Dynasor-CoT & 30 & \textbf{79} & \textbf{1} & {\scriptsize $-$10\%} & 49 & \textbf{87} & \textbf{3} & {\scriptsize $-$13\%} & 39.9 & \textbf{82.7} & \textbf{2.1} & {\scriptsize $-$11\%} \\
AbsBench prompt & 50 & 67 & 4 & {\scriptsize $-$29\%} & 55 & 84 & \underline{5} & {\scriptsize $-$3\%} & 52.2 & 75.9 & 4.6 & {\scriptsize $-$23\%} \\
SES & 60 & 76 & 5 & {\scriptsize $-$30\%} & \textbf{66} & 84 & \underline{5} & {\scriptsize $-$47\%} & \underline{62.7} & 80.3 & 5.0 & {\scriptsize $-$34\%} \\
AtU & \textbf{71} & 76 & 4 & {\scriptsize\underline{$-$41\%}} & \underline{64} & 85 & \underline{5} & {\scriptsize\underline{$-$50\%}} & \textbf{67.6} & 80.7 & 4.5 & {\scriptsize\underline{$-$43\%}} \\
\rowcolor{somabg}\soma{} (ours) & \underline{70} & 75 & 7 & {\scriptsize\boldmath\textbf{$-$44\%}} & 49 & \underline{86} & \textbf{3} & {\scriptsize\boldmath\textbf{$-$59\%}} & 59.4 & 80.7 & 4.8 & {\scriptsize\boldmath\textbf{$-$48\%}} \\
\bottomrule
\end{tabular}
\caption{Unanswerable-question results. Abs is the abstention rate on unanswerable questions, Acc the accuracy on answerable questions and FA the false-abstention rate on answerable questions (all \%); Tok is as in Table~\ref{tab:overthinking}. For \soma{}, the accounting follows the readout in Appendix~\ref{app:readout}. Avg weights SUM and UMWP by their number of questions and is computed from unrounded per-benchmark rates. Bold marks the best value for each S2 model (lowest for FA, largest reduction for Tok); underline marks the second best.}
\label{tab:unanswerable}
\end{table*}

\begin{table}[t]
\centering\footnotesize
\setlength{\tabcolsep}{3pt}
\begin{tabular}{@{}lrrrrrr@{}}
\toprule
 & \multicolumn{2}{c}{Overthinking} & \multicolumn{4}{c}{Unanswerable} \\
\cmidrule(lr){2-3}\cmidrule(lr){4-7}
Variant & Acc$\uparrow$ & Tok & Abs$\uparrow$ & Acc$\uparrow$ & FA$\downarrow$ & Tok \\
\midrule
\multicolumn{7}{@{}l}{\textit{Qwen3-8B}} \\
Vanilla & 93.0 & 6,745 & 72 & 90 & 3 & 3,494 \\
\rowcolor{somabg}\soma{} & 92.5 & {\scriptsize $-$42\%} & 92 & 83 & 10 & {\scriptsize $-$29\%} \\
\quad Remind-only & 93.4 & {\scriptsize $-$16\%} & 74 & 87 & 4 & {\scriptsize $-$12\%} \\
\quad No reminder & 94.1 & {\scriptsize $-$22\%} & 89 & 84 & 10 & {\scriptsize $-$26\%} \\
\midrule
\multicolumn{7}{@{}l}{\textit{R1-Distill-Llama-8B}} \\
Vanilla & 85.0 & 5,299 & 36 & 84 & 3 & 3,972 \\
\rowcolor{somabg}\soma{} & 85.5 & {\scriptsize $-$15\%} & 60 & 82 & 4 & {\scriptsize $-$48\%} \\
\quad Remind-only & 85.7 & {\scriptsize $-$3\%} & 32 & 84 & 0 & {\scriptsize $-$12\%} \\
\quad No reminder & 85.2 & {\scriptsize $-$11\%} & 61 & 82 & 6 & {\scriptsize $-$47\%} \\
\bottomrule
\end{tabular}
\caption{Ablations on a fixed test subset, reported as question-weighted averages. Remind-only never takes over ($h=\infty$). No reminder accumulates evidence from the first pause and takes over without a reminder, with its own calibrated threshold.}
\label{tab:ablation}
\end{table}

\subsection{Setup}
\label{sec:setup}

\paragraph{Scenarios.}
We evaluate \soma{} on overthinking, where the goal is to cut redundant verification without lowering accuracy, and on unanswerable questions, where S2 should abstain when a problem lacks necessary information or has contradictory conditions and solve it otherwise. Both scenarios use the same S1 model, controller and system prompt, and differ only in their monitors (Table~\ref{tab:monitors}).

\paragraph{Models.}
We use Qwen3-8B \citep{yang2025qwen3} and DeepSeek-R1-Distill-Llama-8B \citep{guo2025deepseek} as S2, served with vLLM \citep{kwon2023efficient} in thinking mode with the recommended sampling settings (temperature 0.6, top-$p$ 0.95, and top-$k$ 20 for Qwen3-8B) and a limit of 32{,}768 generated tokens. Every method samples each question once. S1 is Jev \citep{almeida2026introducing,typesafe2026documentation}, accessed through its API.

\paragraph{Benchmarks.}
For overthinking, we use GSM8K (1{,}319 questions; \citealp{cobbe2021training}), MATH-500 (500; \citealp{hendrycks2021measuring,lightman2024lets}), AIME 2024--2025 (60; \citealp{maa2024aime}), AMC 2023 (40; \citealp{maa2023amc}) and GPQA-Diamond (198; \citealp{rein2024gpqa}). For unanswerable questions, we use the test sets of SUM (284 unanswerable and 284 answerable questions; \citealp{song2025hallucination}) and UMWP (300 and 300; \citealp{sun2024benchmarking}), following \citet{liu2026answering}. Mathematical answers are checked with Math-Verify \citep{kydlicek2025mathverify}, and abstentions are identified with the rule of \citet{liu2026answering}.

\paragraph{Implementation.}
S2 pauses every $X=256$ thinking tokens, continuing to the next line break for at most 128 further tokens. The state passed to S1 is limited to 2{,}048 tokens, and a fixed system prompt shared by both scenarios tells S2 that a monitor may insert reminders (Appendix~\ref{app:impl}).

\paragraph{Calibration.}
Thresholds are calibrated as in \S\ref{sec:calibration} with $\delta=0$ on one easier and one harder set of 100 questions per scenario, disjoint from all test sets (Appendix~\ref{app:calib}). The resulting thresholds are $h=0.25$ (Qwen3-8B) and $h=1.5$ (R1-Distill-Llama-8B) for overthinking, and $h=4$ and $h=3$ for unanswerable questions.

\paragraph{Single-run readout.}
Since a takeover ends generation, we run \soma{} once with reminders but no takeovers and read off the outcome under each threshold $h$ at the first pause where the post-reminder evidence reaches $h$, both for calibration and at test time (Appendix~\ref{app:readout}).

\paragraph{Baselines.}
All methods share the decoding settings, generation limit, data, and scoring above; prompts are method-specific. We report Vanilla (thinking mode). For overthinking, we compare with budget forcing (BF; \citealp{muennighoff2025s1}), DEER \citep{yang2026dynamic} and Dynasor-CoT \citep{fu2025reasoning,fu2025efficiently}. For unanswerable questions, we compare with DEER, Dynasor-CoT, the AbstentionBench prompt \citep{kirichenko2025abstentionbench}, statistical early stopping (SES; \citealp{xie2026statistical}) and Answering the Unanswerable (AtU; \citealp{liu2026answering}), which uses a hidden-state probe trained for each S2 model. Parameters that require tuning are selected with each paper's own procedure on the same calibration sets as \soma{}'s thresholds (Appendix~\ref{app:baselines}).

\paragraph{Metrics.}
For overthinking, we report accuracy and the mean number of S2 tokens (Tok), counting thinking and answer tokens as well as extra tokens a method generates, such as trial answers. For unanswerable questions, we report the abstention rate on unanswerable questions (Abs), the accuracy (Acc) and false-abstention rate (FA) on answerable questions, and Tok. For \soma{}, token accounting follows the readout in Appendix~\ref{app:readout}. Averages (Avg) weight each benchmark by its number of questions.

\subsection{Main Results}
\label{sec:results}

\begin{figure}[t]
\centering
\includegraphics[width=\columnwidth]{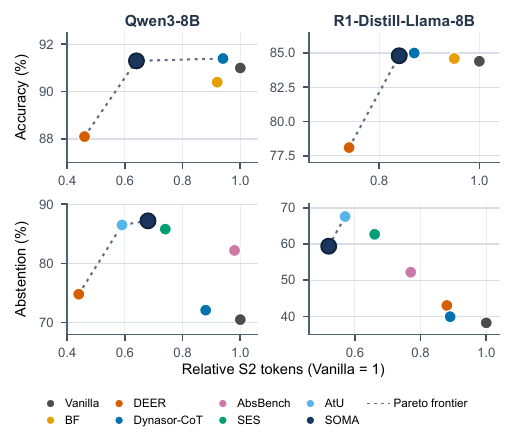}
\caption{Average accuracy on overthinking (top) and abstention on unanswerable questions (bottom) against mean S2 tokens relative to Vanilla, from the Avg columns of Tables~\ref{tab:overthinking} and~\ref{tab:unanswerable}. The dotted line connects non-dominated methods; higher and further left is better. False abstentions are reported in Table~\ref{tab:unanswerable}.}
\label{fig:frontier}
\end{figure}

\paragraph{Overthinking.}
Table~\ref{tab:overthinking} shows that \soma{} shortens reasoning while preserving and slightly improving accuracy on average. It reduces generated tokens by 36\% on Qwen3-8B and 16\% on R1-Distill-Llama-8B, and raises average accuracy from 91.0\% to 91.3\% and from 84.4\% to 84.8\%. Weighting each benchmark equally, which removes the dominance of GSM8K, the gains are 0.8 and 3.4 points, and \soma{} has the highest such average on R1-Distill-Llama-8B (72.0\% vs.\ 69.4\% for Dynasor-CoT). \soma{} exceeds Vanilla on six of the ten model-benchmark combinations, with the largest gains on the hardest benchmarks: on AIME, accuracy rises from 66.7\% to 73.3\% on Qwen3-8B and from 36.7\% to 45.0\% on R1-Distill-Llama-8B, and on GPQA-Diamond it improves by 3.5 and 4.5 points. On the remaining combinations the differences are small, at most 1.4 points except for two questions on AMC 2023 with Qwen3-8B. By contrast, DEER saves more tokens but lowers average accuracy by 2.9 and 6.3 points, and BF stays close to Vanilla in accuracy but saves only 8\% and 5\% of the tokens. Dynasor-CoT keeps accuracy but saves only 6\% of the tokens on Qwen3-8B; on R1-Distill-Llama-8B it saves 13\% with accuracy comparable to \soma{}.

\paragraph{Unanswerable questions.}
Table~\ref{tab:unanswerable} shows that \soma{} substantially improves abstention on unanswerable questions while shortening reasoning. On Qwen3-8B, it raises average abstention from 70.5\% to 87.2\%, the highest among all methods, and reduces tokens by 32\%. On R1-Distill-Llama-8B, it raises abstention from 38.2\% to 59.4\% with the largest token reduction of all methods (48\%). The gains are largest on SUM, the harder benchmark, where abstention rises from 60\% to 87\% and from 28\% to 70\%. On SUM, \soma{} exceeds AtU on Qwen3-8B (87\% vs.\ 83\%) and is one percentage point lower on R1-Distill-Llama-8B (70\% vs.\ 71\%), even though AtU trains a probe on SUM for each S2 model, whereas \soma{} treats S2 as a black box and needs no training. These gains come with a moderate rise in false abstentions on answerable questions, from 2.4\% to 8.0\% and from 2.1\% to 4.8\% on average. On UMWP, \soma{} raises the abstention of Qwen3-8B from 81\% to 88\%. For R1-Distill-Llama-8B, it keeps accuracy and false abstention unchanged while cutting tokens by 59\%, but abstention stays close to Vanilla (49\% vs.\ 48\%). Across both scenarios and both S2 models, \soma{} lies on the token--quality Pareto frontier (Figure~\ref{fig:frontier}), the only method to do so in all four settings.

\subsection{Ablation Study}
\label{sec:ablation}

We study the components of the control layer on a fixed subset of the test sets (Table~\ref{tab:ablation}); Appendix~\ref{app:interval} varies the scan interval. All variants share the S1 model, monitors and system prompt.

\paragraph{Reminders and takeovers.}
\emph{Remind-only} inserts reminders but never takes over ($h=\infty$). In overthinking, reminders alone already shorten reasoning (by 16\% on Qwen3-8B and 3\% on R1-Distill-Llama-8B) without lowering accuracy, and adding takeovers brings total savings to 42\% and 15\%. For unanswerable questions, reminders alone leave abstention close to Vanilla, so the gain in abstention comes from the takeover. \emph{No reminder} skips the first stage: it accumulates evidence from the first pause and takes over without reminding S2, using its own calibrated threshold. For unanswerable questions it performs similarly to \soma{}. In overthinking, the threshold calibrated without a reminder is much higher, and the variant saves about half as many tokens on Qwen3-8B (22\% vs.\ 42\%) with accuracy 1.6 percentage points higher; on R1-Distill-Llama-8B it saves 11\% against 15\% at similar accuracy.

\paragraph{Case studies.}
Appendix~\ref{app:case-studies} shows both stages at work. The typical success is a model that notices a problem but does not act on it: on a SUM question whose denominator has no last term, Qwen3-8B names the gap at once, yet without \soma{} it answers under an assumption after 6{,}148 tokens; under \soma{}, S1 keeps observing the same search after the reminder, and the takeover turns S2's own doubt into an abstention after 1{,}392 tokens. In overthinking, a takeover keeps a confirmed answer, cutting 21{,}665 tokens to 1{,}120. The reminder alone can also suffice: R1-Distill-Llama-8B guesses a missing number on a SUM question without \soma{}, but after the reminder it re-checks the problem and abstains (4{,}903 to 1{,}149 tokens).

\section{Conclusion}
\label{sec:conclusion}

We introduced \soma{}, in which an S1 decision model continuously monitors S2's reasoning for a behavior described in natural language, reminds S2 when it appears, and takes over only if the behavior persists after the reminder. \soma{} requires neither training nor access to S2's internal states, since S1 reads only the visible reasoning text. On overthinking, \soma{} cut reasoning tokens by 36\% and 16\% on the two S2 models without lowering average accuracy, and on unanswerable questions it raised abstention by 17 and 21 points.

Because objectives are written in natural language and S1 reads only visible text, adding a new objective requires only a monitor statement, a reminder and a takeover action, with no change to S2. S1 decision models can therefore serve as a general supervision layer that accompanies reasoning rather than gating it, and in future work the same layer can take on new objectives such as jailbreak defense or agent monitoring.

\FloatBarrier

\clearpage % start Limitations on a new page
% Required by ARR: counts outside the page limit, placed after the conclusion.
\section*{Limitations}

\paragraph{Scope of the evaluation.}
\soma{} is designed as a general framework: a new objective needs only a monitor statement, a reminder and a takeover action, so the same layers could address other problems that surface in the reasoning, such as defending against jailbreaks. Due to time constraints, we evaluate only two scenarios, overthinking and unanswerable questions, both centered on mathematical and scientific reasoning, with two 8B reasoning models and one monitor per scenario. How well \soma{} works in other scenarios and domains, with larger S2 models, and with several monitors running jointly remains to be tested.

\paragraph{A single S1 model.}
All experiments use Jev, a hosted proprietary model whose behavior may change across versions. The framework only requires the S1 interface, but we test no other S1 model in the main experiments. The results therefore do not establish which other S1 models can fill the role.

\paragraph{Cost of S1 calls.}
Jev is accessed through a paid API, and \soma{} queries it at every pause. Our results come from the single-run readout (Appendix~\ref{app:readout}); reported tokens cover S2 only, and we do not measure wall-clock latency. As mature open-source S1 models become available, S1 could be deployed on the same hardware as S2, removing the API cost and the network round trip.

\paragraph{Irreversible interventions.}
A takeover cannot be undone by S2. Although thresholds are chosen to limit harm on calibration data, \soma{} raises false abstentions on answerable questions, and a mistaken takeover in overthinking fixes an answer that S2 might still have corrected.

% Optional, also outside the page limit.
\section*{Broader Impact}

\soma{} lowers the cost of supervising a reasoning model: an objective is written as a natural-language statement rather than a task-specific signal or a trained probe, and S1 reads only the visible reasoning. This could make oversight of long reasoning and agent trajectories practical for users without access to model internals, including for safety objectives such as detecting jailbreak attempts. The same mechanism carries risks. A takeover overrides the model's own reasoning, so a poorly written statement or a badly calibrated threshold can systematically suppress valid answers. Relying on a hosted S1 also sends the model's reasoning to a third-party service, which matters when the reasoning contains private data; a locally deployed open-source S1 would avoid this.

\bibliography{custom}

\appendix
\section{Implementation Details}
\label{app:impl}

\subsection{Models and Decoding}

The S2 models are Qwen3-8B~\citep{yang2025qwen3} and DeepSeek-R1-Distill-Llama-8B~\citep{guo2025deepseek}, and Jev \citep{almeida2026introducing,typesafe2026documentation} is the only S1 model. Every method samples each question once with a generation limit of 32{,}768 tokens and the recommended decoding settings of each model: temperature 0.6 and top-$p$ 0.95, with top-$k$ 20 for Qwen3-8B. Mathematical answers are scored with Math-Verify \citep{kydlicek2025mathverify}, GPQA-Diamond uses its official answer labels, SUM and UMWP use their official answerability labels, and abstentions are identified with the rule of \citet{liu2026answering}. \soma{} and its variants use the fixed system prompt shared by both scenarios:

``While you think, a monitor may insert a \texttt{<reminder>}. When you see one, follow its instruction briefly inside your reasoning, then continue as you judge best.''

\subsection{S1 State and Scan Schedule}

At each scan, Jev receives the complete task and the S2 text in fields named ``Task: $Q$'' and ``Reasoning so far: $H$'', separated by blank lines. The state is limited to 2{,}048 tokens under the Qwen3 tokenizer, after reserving room for the longest monitor statement. The complete task is retained, and the oldest part of $H$ is removed when the limit is exceeded. \soma{} scans only while S2 is thinking. After every 256 thinking tokens, S2 continues to the next line break and pauses there; if no line break appears within 128 further tokens, it pauses without one. Reminders are inserted at the pause as a separate paragraph wrapped in \texttt{<reminder>} tags. They become part of S2's context but are excluded from $H$ on later scans.

\subsection{Takeovers}

In overthinking, a takeover appends the model's end-of-thinking delimiter and ``**Final Answer**'', and S2 then writes at most 64 tokens. For an unanswerable question, it inserts the fixed conclusion ``The problem is missing necessary information or has contradictory conditions, so it cannot be answered as stated.'', closes thinking, pre-fills ``\textbackslash boxed\{I don't know.\}'', and allows S2 512 tokens for the reason.

\section{Single-Run Readout}
\label{app:readout}

Since a takeover ends generation, running every candidate threshold separately would multiply the cost of calibration by the size of the grid. We therefore run \soma{} once in an \emph{observe-remind} mode: reminders are inserted as usual, but no takeover is executed, and every scan is recorded. A trial takeover is recorded whenever the post-reminder $S$ first reaches a candidate threshold. The outcome under a threshold $h$ is the trial at the first pause where $S\ge h$, or the full trajectory if $h$ is never reached; $h=\infty$ gives the remind-only variant. Up to the first reminder, \soma{} coincides with S2 running under the same system prompt without intervention, so each such trajectory is generated once and \soma{} continues from it. The same readout, with the frozen thresholds, is used for evaluation. At the pause where the takeover would have happened, the outcome and token count are computed as follows.
\begin{itemize}
\item \textbf{Overthinking.} We append the takeover text and let S2 write its final answer (at most 64 tokens). The answer is scored as usual. The token count is the number of tokens S2 wrote before the pause, plus the takeover text and the final answer.
\item \textbf{Unanswerable questions.} The pre-filled refusal determines the outcome, so the trial is scored as an abstention. S2 then writes the reason (at most 512 tokens), and the token count is the number of tokens S2 wrote before the pause, plus the tokens of the generated reason.
\end{itemize}

\section{Calibration and Ablation Data}
\label{app:calib}

The calibration sets are disjoint from all test questions, and each scenario has an easier and a harder set of 100 questions. For overthinking, the easier set consists of MATH training problems sampled to match the difficulty distribution of MATH-500, and the harder set of problems from AIME 2020--2023. For unanswerable questions, the easier set contains 50 answerable and 50 unanswerable problems drawn from the original UMWP release but excluded from its test split, and the harder set contains 50 SUM training pairs, each consisting of an answerable problem and its unanswerable counterpart. We search $\mathcal{H}=\{0.25,0.5,0.75,1,1.5,2,2.5,3,4,6,8,12\}$ with $\delta=0$. Table~\ref{tab:calibration} reports the calibration outcomes at the selected thresholds of \soma{} and of the No-reminder variant.

\begin{table}[t]
\centering\small
\setlength{\tabcolsep}{4pt}
\begin{tabular}{@{}llrrr@{}}
\toprule
S2 & Variant & $h$ & Takeovers & $+$/$-$ \\
\midrule
\multicolumn{5}{@{}l}{\textit{Overthinking}} \\
Qwen3 & \soma{} & 0.25 & 161 & $+$3/$-$1 \\
Qwen3 & No reminder & 6 & 101 & $+$1/$-$0 \\
Llama-R1 & \soma{} & 1.5 & 70 & $+$3/$-$2 \\
Llama-R1 & No reminder & 3 & 74 & $+$0/$-$0 \\
\midrule
\multicolumn{5}{@{}l}{\textit{Unanswerable}} \\
Qwen3 & \soma{} & 4 & 30 & $+$18/$-$3 \\
Qwen3 & No reminder & 8 & 33 & $+$18/$-$2 \\
Llama-R1 & \soma{} & 3 & 38 & $+$31/$-$1 \\
Llama-R1 & No reminder & 2.5 & 42 & $+$32/$-$1 \\
\bottomrule
\end{tabular}
\caption{Calibration outcomes at the selected thresholds, pooled over the two calibration sets of each scenario (200 questions per scenario). Takeovers counts the questions taken over; $+$/$-$ counts questions turned from wrong to correct and from correct to wrong relative to S2 under the same system prompt without intervention.}
\label{tab:calibration}
\end{table}

The ablations in \S\ref{sec:ablation} use a fixed subset of the test sets, selected by hashing question identifiers: all of MATH-500 (500 questions) and AIME (60) for overthinking, and 100 SUM pairs and 100 answerable plus 100 unanswerable UMWP questions for unanswerable questions.

\section{Scan Interval}
\label{app:interval}

\begin{table}[t]
\centering\footnotesize
\setlength{\tabcolsep}{3pt}
\begin{tabular}{@{}lrrrrrr@{}}
\toprule
 & \multicolumn{2}{c}{Overthinking} & \multicolumn{4}{c}{Unanswerable} \\
\cmidrule(lr){2-3}\cmidrule(lr){4-7}
Variant & Acc$\uparrow$ & Tok & Abs$\uparrow$ & Acc$\uparrow$ & FA$\downarrow$ & Tok \\
\midrule
\multicolumn{7}{@{}l}{\textit{Qwen3-8B}} \\
Vanilla & 93.0 & 6,745 & 72 & 90 & 3 & 3,494 \\
$X{=}128$ & 92.9 & {\scriptsize $-$48\%} & 93 & 85 & 10 & {\scriptsize $-$36\%} \\
\rowcolor{somabg}$X{=}256$ & 92.5 & {\scriptsize $-$42\%} & 92 & 83 & 10 & {\scriptsize $-$29\%} \\
$X{=}512$ & 94.1 & {\scriptsize $-$34\%} & 88 & 86 & 6 & {\scriptsize $-$27\%} \\
\midrule
\multicolumn{7}{@{}l}{\textit{R1-Distill-Llama-8B}} \\
Vanilla & 85.0 & 5,299 & 36 & 84 & 3 & 3,972 \\
$X{=}128$ & 84.8 & {\scriptsize $-$19\%} & 63 & 83 & 4 & {\scriptsize $-$51\%} \\
\rowcolor{somabg}$X{=}256$ & 85.5 & {\scriptsize $-$15\%} & 60 & 82 & 4 & {\scriptsize $-$48\%} \\
$X{=}512$ & 84.8 & {\scriptsize $-$8\%} & 55 & 82 & 4 & {\scriptsize $-$43\%} \\
\bottomrule
\end{tabular}
\caption{\soma{} with different scan intervals $X$ (thinking tokens) on the ablation subset of Table~\ref{tab:ablation}. $X{=}256$ is the main setting; the other intervals reuse its thresholds.}
\label{tab:interval}
\end{table}

Scanning more often lets \soma{} notice a behavior earlier (Table~\ref{tab:interval}). Halving the interval to $X=128$ increases savings in both scenarios (to 48\% and 19\% in overthinking, and to 36\% and 51\% for unanswerable questions) and raises abstention on R1-Distill-Llama-8B from 60\% to 63\%, at the cost of about 1.5--2 times as many S1 calls. Doubling it to $X=512$ roughly halves the S1 calls but reduces savings and, on R1-Distill-Llama-8B, abstention, while slightly raising overthinking accuracy on Qwen3-8B through fewer takeovers. The interval thus trades monitoring cost against responsiveness, and $X=256$ lies between the two.

\section{Baselines}
\label{app:baselines}

We reproduce all baselines from their released code, following the specifications reported in their papers. Where a method has parameters that must be tuned, we select them with the procedure of its paper on the same calibration sets used for \soma{}'s thresholds (Appendix~\ref{app:calib}), separately for each S2 model and scenario. All baselines share the task prompts, decoding settings, data and generation limit of \S\ref{sec:setup}, while keeping their own intervention prompts.

\paragraph{Budget forcing (BF) \citep{muennighoff2025s1}.} A length-control method that caps thinking at a fixed token budget: once the budget is reached, thinking is cut off and the model is prompted to give its final answer. The budget is selected on the calibration sets.

\paragraph{DEER \citep{yang2026dynamic}.} An early-exit method based on the reasoner's own confidence. At reasoning transitions such as ``Wait'', the reasoner writes a trial answer, and thinking ends once its confidence in that answer is high enough, using the threshold of the paper.

\paragraph{Dynasor-CoT \citep{fu2025reasoning,fu2025efficiently}.} An early-exit method based on answer consistency. At fixed intervals, an intermediate answer is elicited and then discarded, and the method exits once several consecutive answers agree and show no hesitation. The probing interval and the number of consecutive answers are selected on the calibration sets.

\paragraph{Answering the Unanswerable \citep{liu2026answering}.} A white-box method that monitors S2's hidden states with a linear probe trained for each model. When the probe indicates that the question is unanswerable, it inserts guidance that leads S2 to abstain. We use the released probe weights and select the probe's decision threshold on the calibration sets.

\paragraph{Statistical early stopping \citep{xie2026statistical}.} A method that tracks how often uncertainty keywords appear in the reasoning and stops when a statistical test finds that they appear more often than is typical for answerable questions. Its statistics for answerable questions are estimated on the answerable questions of the calibration sets, and the remaining settings follow the paper. The paper proposes a renewal and a maxwise stopping rule; we report the renewal rule, the stronger of the two in our runs. Because the method does not specify an action after stopping, it uses our refusal takeover.

\paragraph{AbstentionBench prompt \citep{kirichenko2025abstentionbench}.} A prompt-only baseline that adds the official AbstentionBench system prompt, which encourages the model to abstain when a question cannot be answered.

\newtcolorbox{casebox}[1]{breakable, enhanced, colback=black!2, colframe=black!45,
  boxrule=0.5pt, arc=1pt, left=5pt, right=5pt, top=3pt, bottom=3pt,
  fonttitle=\bfseries\small, title={#1}, fontupper=\small}
\newcommand{\excerpt}[1]{\begin{list}{}{\leftmargin=1.2em\rightmargin=0.4em\topsep=2pt}\item\relax\itshape #1\end{list}}
\newcommand{\inserted}[1]{\par\smallskip\noindent\colorbox{blue!7}{\parbox{\dimexpr\linewidth-2\fboxsep\relax}{\ttfamily\footnotesize #1}}\par\smallskip}
\newcommand{\caselabel}[1]{\par\smallskip\noindent\textbf{#1}\ }

\section{Case Studies}
\label{app:case-studies}

This appendix shows how SOMA behaves on individual questions, with four successful and two failed cases. All cases come from the main test run. For each case we show the question, the vanilla response, and SOMA's monitoring trace. At every scan the trace lists the position in S2 tokens, S1's probability $p$ that the monitor statement holds, the evidence after the update, and the action.

\begin{table*}[t]
\centering
\small
\setlength{\tabcolsep}{4pt}
\begin{tabular}{lllllrr}
\toprule
Case & Scenario & S2 & Decisive step & Outcome (vanilla $\rightarrow$ SOMA) & Vanilla & SOMA \\
\midrule
1 & UMWP & R1 & takeover & no answer $\rightarrow$ abstains & 32{,}768 & 1{,}245 \\
2 & SUM & Qwen & takeover & answers under an assumption $\rightarrow$ abstains & 6{,}148 & 1{,}392 \\
3 & SUM & R1 & reminder & guesses the missing number $\rightarrow$ abstains & 4{,}903 & 1{,}149 \\
4 & MATH-500 & R1 & takeover & correct $\rightarrow$ correct & 21{,}665 & 1{,}120 \\
\midrule
5 & MATH-500 & R1 & takeover & correct $\rightarrow$ wrong intermediate answer & 1{,}679 & 3{,}147 \\
6 & SUM & Qwen & takeover & correct $\rightarrow$ false refusal & 5{,}544 & 2{,}436 \\
\bottomrule
\end{tabular}
\caption{Cases in this appendix. The last two columns give the tokens generated by S2.}
\label{tab:case-overview}
\end{table*}

\subsection{Successful cases}

\begin{casebox}{Case 1 \textbullet{} UMWP, R1: a takeover ends an assumption loop}
\caselabel{Question.} A pole is 20 meters long. It got cut for many pieces in a way that left it 30\% shorter. How long is the maximum piece of pole?

\caselabel{Vanilla} (32{,}768 tokens, stops at the generation limit without an answer).
\excerpt{Calculating 30\% of 20 meters: $0.3 \times 20 = 6$ meters. So, the total length after cutting is $20 - 6 = 14$ meters. [\dots] Wait, but the problem says it's cut into many pieces [\dots]\\
\emph{(near the end)} I think I need to conclude that the maximum possible length is approaching 14 meters, but since it's expecting a specific answer, maybe 10 meters. But I'm not sure. Wait, maybe the problem is expecting 10 meters, but I don't have a logical reason.}
The response contains 694 occurrences of ``Wait'' and never settles.

\caselabel{SOMA trace.} At token 283, S1 gives $p=0.22$ and $S$ stays $0$. At token 554, $p=0.75$ makes $S$ positive ($1.10$): the reminder is inserted and $S$ is reset to $0$. The next two scans give $p=0.87$ and $p=0.90$, raising $S$ to $1.90$ and then to $4.10 \geq h=3$, and the takeover fires at token 1{,}189.

\caselabel{Before the reminder.}
\excerpt{Wait, maybe the question is asking for the maximum possible length regardless of the number of pieces, so theoretically, the maximum possible is 14 meters. But since it's cut into many pieces, maybe it's implying that it's split into more than one, but not necessarily a specific number.}
\caselabel{After the reminder.}
\excerpt{Wait, the problem says it was cut into many pieces, leaving it 30\% shorter. [\dots] The term ``many'' is a bit vague, but in math problems, often ``many'' is used to imply two or more. So, if it's cut into two pieces, each would be 7 meters.}
\caselabel{Takeover output} (token 1{,}189).
\inserted{The problem is missing necessary information or has contradictory conditions, so it cannot be answered as stated.\\ \textless/think\textgreater\\ \textbackslash boxed\{I don't know.\}\\ Reason\{The problem does not specify the number of pieces, making it impossible to determine the maximum length of any single piece.\}}
\caselabel{Outcome.} Correct abstention after 1{,}245 tokens instead of an unfinished 32{,}768-token response.
\end{casebox}

\noindent The S2 model notices the ambiguity before the reminder but keeps resolving it by assumption. The check reminder does not change this: S2 decides that ``many'' means two pieces. S1 keeps observing the same behaviour ($p=0.87$, $0.90$), the post-reminder evidence reaches $h$ after two scans, and the takeover ends the episode. The reason is written by S2 and names the missing quantity.

\begin{casebox}{Case 2 \textbullet{} SUM, Qwen3-8B: noticed at the start, acted on only through the takeover}
\caselabel{Question.} $\dfrac{2+4+6+\cdots+34}{3+6+9+\cdots} = {?}$ \quad (the last term of the denominator is missing)

\caselabel{Vanilla} (6{,}148 tokens, answers $\frac{2}{3}$).
\excerpt{Wait, actually, the denominator is written as $3 + 6 + 9 + \cdots$ but it doesn't specify the last term. Hmm, that might be an issue. Let me check again. [\dots]\\
\emph{(final answer)} Based on the assumption that both the numerator and denominator sequences have the same number of terms (17), the final result is: $\boxed{\frac{2}{3}}$}

\caselabel{SOMA trace.} At the first scan (token 384), $p=0.63$ makes $S$ positive ($0.53$): the reminder is inserted and $S$ is reset to $0$. The next three scans (tokens 645, 925 and 1{,}309) give $p=0.85$, $0.88$ and $0.94$, raising $S$ to $1.73$, $3.73$ and $6.48 \geq h=4$, and the takeover fires at token 1{,}309.

\caselabel{After the reminder.}
\excerpt{Now the denominator is $3 + 6 + 9 + \cdots$ but what's the last term? The problem didn't specify. Wait, maybe it's supposed to end at the same term as the numerator? [\dots] If they are supposed to have the same number of terms? Or maybe the denominator ends at the same value as the numerator?}
\caselabel{Takeover output} (token 1{,}309).
\inserted{The problem is missing necessary information or has contradictory conditions, so it cannot be answered as stated.\\ \textless/think\textgreater\\ \textbackslash boxed\{I don't know.\}\\ Reason\{The problem does not specify the last term of the denominator sequence (3 + 6 + 9 + ...). Without knowing the endpoint of the denominator, it is impossible to compute its sum and thus determine the value of the given fraction.\}}
\caselabel{Outcome.} Correct abstention after 1{,}392 tokens; vanilla spends 6{,}148 tokens and answers under an explicit assumption.
\end{casebox}

\noindent This is the ``knowing but not acting'' pattern that motivates SOMA: vanilla names the missing endpoint in its first paragraph and still produces a definite answer. With SOMA, S2 again searches for an implied endpoint after the reminder; S1's evidence grows on every scan, and the takeover turns the doubt S2 already expressed into an abstention.

\begin{casebox}{Case 3 \textbullet{} SUM, R1: the reminder alone leads to an abstention}
\caselabel{Question.} What is the value of $\dfrac{(2112-\ )^2}{169}$?

\caselabel{Vanilla} (4{,}903 tokens, answers 26244).
\excerpt{I wonder what the missing part is in the numerator. It seems like there's a blank space where a number should be. [\dots]\\
\emph{(final answer)} Substitute $x = 6$ into the numerator: $2112 - 6 = 2106$. [\dots] Thus, the value of the given expression is $\boxed{26244}$.}

\caselabel{SOMA trace.} The first two scans (tokens 276 and 599) give $p=0.49$ and $p=0.35$, so $S$ stays $0$. At token 904, $p=0.67$ makes $S$ positive ($0.71$): the reminder is inserted and $S$ is reset to $0$. S2 ends its reasoning before the next scan, so no takeover occurs.

\caselabel{Before the reminder.}
\excerpt{Wait, perhaps the missing number is 12? Because 2112 minus 12 is 2100, and 2100 squared is a big number, but 2100 divided by 169 is approximately 12.36, which isn't an integer.}
\caselabel{After the reminder.}
\excerpt{After checking again, the problem still lacks information about the missing number in the numerator. Therefore, I can't solve it without knowing what's being subtracted from 2112. So, I think the correct response here is to state that the problem lacks sufficient information and that I don't know the solution. \textbf{Final Answer} $\boxed{\text{I don't know.}}$}
\caselabel{Outcome.} Correct abstention after 1{,}149 tokens, without a takeover.
\end{casebox}

\noindent Here the check reminder does what it asks for: S2 re-examines the problem, concludes that information is missing, and ends its reasoning by itself. Such cases are the minority in our runs. On the full SUM test set the reminder alone raises R1's abstention rate only from 28\% (Vanilla) to 30\%, and most of SOMA's gain comes from the takeover.

\begin{casebox}{Case 4 \textbullet{} MATH-500, R1: takeover after a confirmed answer}
\caselabel{Question.} Simplify $\dfrac{\sec x}{\sin x} - \dfrac{\sin x}{\cos x}$.

\caselabel{Vanilla} (21{,}665 tokens, correct: $\cot x$).
\excerpt{\emph{(after reaching $\cot x$ and re-verifying it several times, a numerical check at $x=\pi/6$ with an arithmetic slip)} $\dfrac{\sec \frac{\pi}{6}}{\sin \frac{\pi}{6}} - \dfrac{\sin \frac{\pi}{6}}{\cos \frac{\pi}{6}} = \dfrac{2}{\frac{1}{2}} - \dfrac{\frac{1}{2}}{\frac{\sqrt{3}}{2}} = 4 - \dfrac{1}{\sqrt{3}}$ [\dots] So, original expression is $\frac{12 - \sqrt{3}}{3}$, not $\cot x$. Wait, so my algebra is wrong.}
The slip ($\sec\frac{\pi}{6}$ is $\frac{2}{\sqrt{3}}$, not 2) starts a long re-derivation; the response contains 242 occurrences of ``Wait'' before it answers $\boxed{\cot x}$.

\caselabel{SOMA trace.} At token 333, S1 gives $p=0.03$. At token 717, $p=0.66$ makes $S$ positive ($0.66$): the reminder is inserted and $S$ is reset to $0$. At the next scan (token 1{,}101), $p=0.91$ gives $S=2.31 \geq h=1.5$, and the takeover fires.

\caselabel{Before the reminder.}
\excerpt{But wait, $\frac{\cos x}{\sin x}$ is the definition of $\cot x$, right? So, does that mean the simplified expression is $\cot x$? Hmm, let me double-check my steps to make sure I didn't make a mistake.}
\caselabel{After the reminder.}
\excerpt{Okay, so after verifying, I'm confident that the steps are correct. So, the simplified form is indeed $\cot x$. [\dots] Let me just make sure there isn't another way to approach this problem that might lead to a different form, but I don't think so.}
\caselabel{Takeover output} (token 1{,}101).
\inserted{\textless/think\textgreater\\ **Final Answer**\\ The simplified expression is \textbackslash(\textbackslash boxed\{\textbackslash cot x\}\textbackslash).}
\caselabel{Outcome.} Correct after 1{,}120 tokens instead of 21{,}665 ($-95\%$).
\end{casebox}

\noindent After the reminder, S2 states that its checks have confirmed the answer, yet it starts another round (``Let me just make sure \dots''). This is exactly the observable symptom of the monitor statement. S1 scores it with $p=0.91$ at the next scan, and the takeover keeps the confirmed answer. Vanilla's extra verification adds no information; its numerical check even introduces a spurious doubt.

% ---------------------------------------------------------------------
\subsection{Failure cases}

\begin{casebox}{Case 5 \textbullet{} MATH-500, R1: the takeover locks in an error that the next check would have caught}
\caselabel{Question.} Consider the geometric sequence $\frac{125}{9}, \frac{25}{3}, 5, 3, \ldots$. What is the eighth term of the sequence? Express your answer as a common fraction.

\caselabel{Vanilla} (1{,}679 tokens, correct: $\frac{243}{625}$).

\caselabel{SOMA run before the reminder} (an arithmetic slip while simplifying $\frac{273{,}375}{703{,}125}$: the numerator is divided by 3, the denominator is not).
\excerpt{$30{,}375 \div 3$: [\dots] So, $10{,}125$. $78{,}125$: Since it's not divisible by 3, we can't reduce further. So, the fraction is $10{,}125 / 78{,}125$. [\dots] So, $81/625$. Is this reducible? [\dots] So, $81/625$ is the simplified form.}

\caselabel{SOMA trace.} For the first eight scans (tokens 333 to 2{,}585), S1 gives $p \leq 0.08$. At token 2{,}842, after S2 has written $81/625$, $p=0.62$ makes $S$ positive ($0.49$): the reminder is inserted and $S$ is reset to $0$. At the next scan (token 3{,}122), $p=0.85$ gives $S=1.73 \geq h=1.5$, and the takeover fires. Without the takeover, the following scan (token 3{,}399) would have given $p=0.19$, and $S$ would have fallen to $0.28$.

\caselabel{After the reminder.}
\excerpt{Wait, so I have $81/625$ as the simplified fraction. Let me confirm all the steps to make sure I didn't make a mistake. [\dots] Reduced the fraction step by step, ending up with $81/625$. Wait, let me verify the multiplication again: $125 \times 2187$.}
\caselabel{Takeover output} (token 3{,}122).
\inserted{\textless/think\textgreater\\ **Final Answer**\\ The eighth term of the sequence is \textbackslash boxed\{\textbackslash dfrac\{81\}\{625\}\}.}
\caselabel{Outcome.} Wrong ($\frac{81}{625}$). Without the takeover, the same trajectory notices the inconsistency right away (``Wait, that seems inconsistent. [\dots] So, I must have made [a mistake]''; $p$ drops to 0.19 at the next scan) and ends with the correct $\frac{243}{625}$ after 6{,}402 tokens.
\end{casebox}

\noindent S1 judges the visible text correctly: an answer has been reached and S2 is re-checking it without naming a concern. But S1 does not, by design, judge whether the answer is right, and this particular check would have found the error. This is the residual risk of early exit. The reset makes the takeover wait for evidence after the reminder, and the calibrated $h$ keeps such cases rare.

\begin{casebox}{Case 6 \textbullet{} SUM, Qwen3-8B: a false refusal on a stem that looks incomplete}
\caselabel{Question.} If $3x^3-9x^2+kx-12$ is divisible by $x-3$, then it is also divisible by:

\caselabel{Vanilla} (5{,}544 tokens, correct: $\boxed{3x^2+4}$).

\caselabel{SOMA trace.} The first two scans (tokens 285 and 657) give $p=0.43$ and $p=0.47$, so $S$ stays $0$. At token 944, $p=0.64$ makes $S$ positive ($0.58$): the reminder is inserted and $S$ is reset to $0$. The next five scans (tokens 1{,}253 to 2{,}341) give $p$ between $0.66$ and $0.72$, so $S$ grows to $0.66$, $1.51$, $2.31$, $3.21$ and $4.15 \geq h=4$, and the takeover fires at token 2{,}341.

\caselabel{After the reminder.}
\excerpt{Therefore, the polynomial factors as $(x - 3)(3x^2 + 4)$. [\dots] But if the question is asking for another factor, maybe it's expecting a linear factor? [\dots] maybe the original problem had options like $x - 3$, $x + 3$, $3x^2 + 4$, etc. But since the user hasn't provided options, I need to think again.}
\caselabel{Takeover output} (token 2{,}341).
\inserted{The problem is missing necessary information or has contradictory conditions, so it cannot be answered as stated.\\ \textless/think\textgreater\\ \textbackslash boxed\{I don't know.\}\\ Reason\{The problem as stated does not provide sufficient information to determine the additional factor. While the value of \$ k \$ can be determined using the Factor Theorem, the question asks for another divisor without specifying options or further conditions. Without additional context or choices, it is not possible to definitively identify the other factor.\}}
\caselabel{Outcome.} False refusal; vanilla answers correctly.
\end{casebox}

\noindent S2 has already found the intended factor, but it keeps questioning whether answer options are missing. The monitor statement describes this visible behavior truthfully, and moderate probabilities ($p=0.64$ to $0.72$) accumulate over five scans until they reach $h=4$. S1 judges what S2 does, not whether the information S2 misses is really necessary. False refusals therefore arise when S2 voices unwarranted doubts.

\paragraph{Summary.} In the successful cases, S1 detects a persistent, observable symptom: an assumption made despite missing information, or verification of an answer that is already confirmed. The reminder occasionally resolves the case by itself; otherwise, the evidence S1 collects after the reminder triggers the takeover. The failures have two distinct causes. The first is a check that would have been productive, which S1 cannot recognize without judging correctness. The second is S2 voicing doubts that S1 correctly observes but that are unwarranted.

\end{document}